\documentclass[twoside]{article}

\usepackage{graphicx}
\usepackage{tabularx}
\usepackage{adjustbox}
\usepackage{pifont}
\usepackage{algorithm}
\usepackage{algpseudocode}
\usepackage{enumitem}
\usepackage{amsmath}
\usepackage{xcolor}
\usepackage{subcaption}
\usepackage{arydshln}
\usepackage{amssymb}
\usepackage{url}
\usepackage[accepted]{claiunconf2023}

\usepackage[round]{natbib}

\begin{document}

%

%

\twocolumn[

\CLAIUnconftitle{AdaCL: Adaptive Continual Learning}

\CLAIUnconfauthor{ Elif Ceren Gok Yildirim \And Murat Onur Yildirim \And  Mert Kilickaya \And Joaquin Vanschoren }

\CLAIUnconfaddress{ Automated Machine Learning Group, Eindhoven University of Technology } ]

\begin{abstract}
Class-Incremental Learning aims to update a deep classifier to learn new categories while maintaining or improving its accuracy on previously observed classes. Common methods to prevent forgetting previously learned classes include regularizing the neural network updates and storing exemplars in memory, which come with hyperparameters such as the learning rate, regularization strength, or the number of exemplars. However, these hyperparameters are usually only tuned at the start and then kept fixed throughout the learning sessions, ignoring the fact that newly encountered tasks may have varying levels of novelty or difficulty. This study investigates the necessity of hyperparameter `adaptivity' in Class-Incremental  Learning: the ability to dynamically adjust hyperparameters such as the learning rate, regularization strength, and memory size according to the properties of the new task at hand. We propose AdaCL, a Bayesian Optimization-based approach to automatically and efficiently determine the optimal values for those parameters with each learning task. We show that 
 adapting hyperpararmeters on each new task leads to improvement in accuracy, forgetting and memory. Code is available at \url{https://github.com/ElifCerenGokYildirim/AdaCL}.

\end{abstract}

\section{Introduction}
This paper focuses on Class-Incremental Learning of deep neural network representations~\citep{masana2020class,defying}. Unlike standard batch learning, which requires access to data from all categories simultaneously, Class-Incremental Learning can update a pre-trained deep classifier with new categories by expanding the classifier layer with new output nodes for new classes. This leads to more efficient learning and avoids the need to store task identities which often are not available in real-world scenarios.
\vspace{1pt}

While Class-Incremental Learning enables expanding a classifier without requiring task identities, it often results in \emph{catastrophic forgetting}. This occurs when the deep learner sacrifices accuracy on previously seen classes to learn new ones. Three major approaches have been explored to address this issue: regularization, replay and architecture adaptation. Regularization prevents abrupt shifts in the neural network weights while learning new classes~\citep{kirkpatrick2017overcoming,li2017learning}. Replay stores a few exemplars per class in memory and replays them during new learning increments~\citep{lopez2017gradient}. Architecture-based approaches build network structures by either expanding the existing network \citep{pnn, der} or by partially isolating network parameters to retain past class information \citep{aanets, wsn, cps}. Although these methods improve the performance, they always use a fixed learning rate, regularization magnitude, and pre-defined memory size throughout the learning process, which is likely suboptimal.

\begin{figure}[t]
  \centering
  \includegraphics[width=0.5\textwidth]{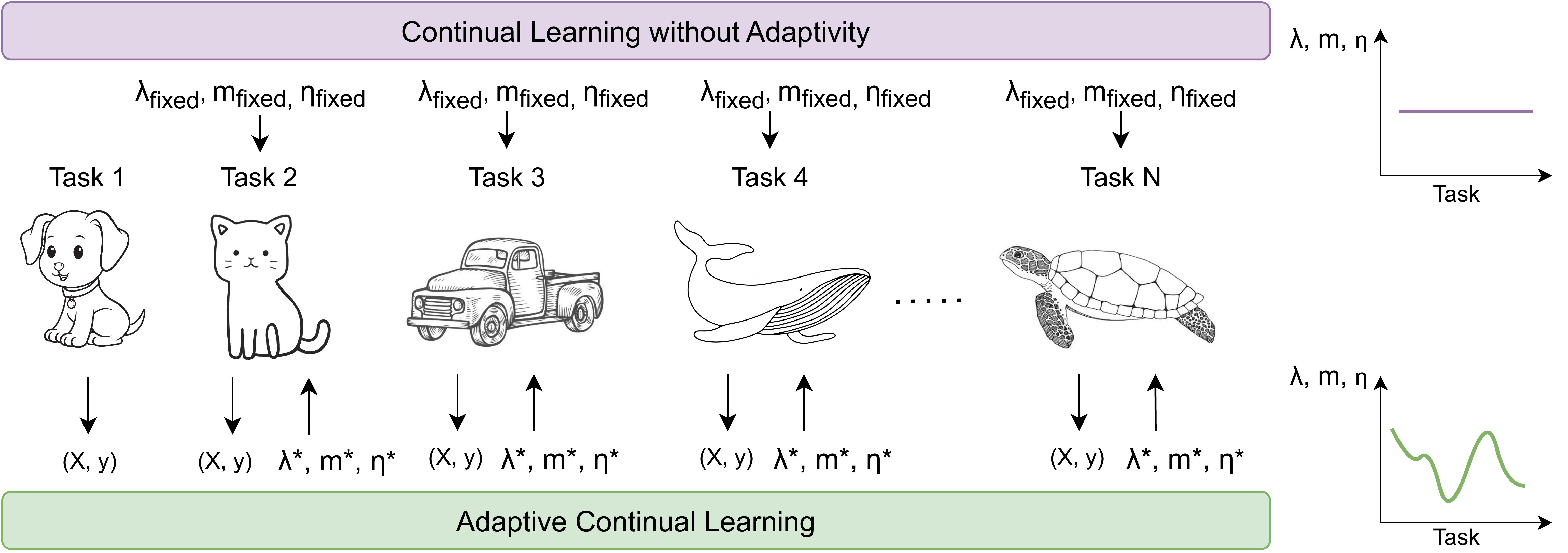}
  \caption{Comparison of fixed \textit{vs.} adaptive continual learning (AdaCL). In this work, we hypothesize that different tasks may require different settings and explore the potential of tuning learning rate ($\eta$), regularization strength ($\lambda$) and memory size per task ($m$), allowing to learn adaptively.}
  \label{fig:teaser}
\vspace{-15pt}
\end{figure}

This paper addresses the issue of dynamically adjusting \emph{how much} to regularize or store in memory for each new task. We explore whether adaptation is necessary for optimal performance, treating the learning rate, regularization magnitude, and memory size as latent variables that should be adjusted based on the current state of the learner and the complexity of the task (see Figure~\ref{fig:teaser}). We use Bayesian Optimization to efficiently discover the best hyperparameters per task. Our experiments on CIFAR-100 and MiniImageNet demonstrate that adapting these parameters to the tasks results in significant improvements and give us new insight into how to adapt to various new tasks. In summary, this paper makes the following contributions: 

\begin{enumerate}[label=\Roman*.]
\item In this paper, for the first time, we raise the important issue of adaptive hyperparameter selection in class-incremental learning.\vspace{5pt}
\item We propose to predict the learning rate, regularization magnitude, and memory size conditioned on the state of the deep learner and the current learning task via Bayesian Optimization.\vspace{5pt}
\item Through large-scale experiments on well-established benchmarks, we show that learning adaptively yields significant performance and efficiency improvements, both increasing accuracy and reducing forgetting. 
\end{enumerate} 

\section{Related Work}
\textbf{Class-Incremental Learning.} Class-Incremental Learning updates a deep classifier with sequentially arriving data, usually with mutually exclusive categories~\citep{masana2020class, defying, wang2023comp, zhou2023class, kilickaya2023towards}. However, when novel data arrives, previous training data becomes unavailable, leading to catastrophic forgetting. To mitigate this, researchers have developed three main approaches: (i) regularization-based methods, which stabilize important parameters or distill previous knowledge into the model~\citep{kirkpatrick2017overcoming,  zenke2017continual, lee2017overcoming, li2017learning, chaudhry2018riemannian, zhou2021co, pass}, (ii) replay-based methods, which usually benefit from regularization-based methods and store a subset of training data to rehearse during learning~\citep{rebuffi2017icarl, chaudhry2018efficient, wu2019large,   aljundi2019gradient, ostapenko2019learning, xiang2019incremental, zhao2020maintaining, rmm, fetril} and (iii) architecture-based methods redesign network architectures by extending the network~\citep{pnn, der, ssre} or freezing network parameters partially to preserve old class knowledge~\citep{aanets, wsn, cps}.

\vspace{5pt}
However, current studies assume a constant amount of regularization and memory size per task throughout learning sessions, which is unnatural since learning unfamiliar objects requires more plasticity than learning familiar ones~\citep{cha2024hyperparameters}. To address this issue, we propose an adaptive method in which the regularization magnitude, learning rate and memory size are automatically tuned within each incremental learning step.

\textbf{Hyperparameter Optimization.} Hyperparameter Optimization (HPO) aims to optimize the hyperparameters of a given deep learning model, including the learning rate, layer size, or balance of different loss functions. In this paper, our focus is on balancing the contribution of a standard cross-entropy and regularization loss, learning rate as well as memory size per task if applicable. To tackle the HPO problem, complex techniques such as bi-level optimization~\citep{franceschi2018bilevel} or gradient-based optimization~\citep{baydin2018automatic} have been proposed. Bi-level optimizers alternate between optimizing neural network weights and tuning the hyper-parameters, while gradient-based methods treat all network weights as hyperparameters to be updated.

\vspace{5pt}
Several recent studies~\citep{AGEM, defying, omdp} share our core motivation by investigating the impact of hyperparameter optimization in subsequent tasks. ~\citet{defying} adopt a two-stage strategy: First, they fine-tune the current task to identify the optimal learning rate with a grid search for maximum plasticity and peak accuracy. Second, they introduce a new thresholding hyperparameter to naively balance the plasticity and stability trade-off: starting with a high regularization strength and decaying it when the performance of the current task is below the defined threshold. However, this approach follows a very naive search since they basically apply two consecutive grid searches to decide the optimal value. Moreover, they focus on a Task-Incremental setup and do not consider the memory size in their search space.

\vspace{5pt}
{\citet{AGEM} tunes the hyperparameters for the first $T$ tasks with a grid search and then uses the best-found values in the remaining tasks. However, it assumes that the initial few tasks are representative enough for the rest of the tasks which may not be realistic in most of the cases. Again, they worked on the Task-Incremental scenario and did not consider the memory size in their search space.

\vspace{5pt}
{\citet{omdp} uses reinforcement learning in a Class-Incremental scenario to adaptively find the best hyperparameter values while learning the tasks. They hold a validation set, similar to our study, to estimate rewards by finding the best set of hyperparameters. However, its search space is limited to learning rate, regularization strength, and the type of classifier.

\vspace{5pt}
In this work, we propose Bayesian Optimization~\citep{snoek2012practical} with Tree Parzen Estimators due to its effectiveness over multiple hyperparameters. We evaluate the generality of our approach by dynamically tuning the learning rate, regularization strength, and memory size across a stream of tasks.

\section{Method}
\label{sec:empirical method}
\textbf{Overview.} Class-incremental learning involves updating a neural network with new classes as it comes in. Specifically, the learner receives a sequence of learning tasks $\mathcal{T}_{1:t} = (\mathcal{T}_{1}, \mathcal{T}_{2}, ...,\mathcal{T}_{t})$, each with a corresponding dataset $\mathcal{D}_{\mathcal{T}} = { (x_{i,t}, y_{i,t})^{n_{t}} }$ consisting of $n_{t}$ instances per task. Each input pair ${x_{i,t}, y_{i,t}} \in \mathcal{X}_{t} \times \mathcal{Y}_{t}$ is sampled from an unknown distribution where $x_{i,t}$ is the sample and $y_{i,t}$ is the corresponding label. It's important to note that the learning tasks are mutually exclusive, i.e., $\mathcal{Y}_{t-1} \cap \mathcal{Y}_{t} = \emptyset$. When a new learning task arrives, the deep convolutional network is optimized to embed the input instance into the classifier space $f_{\Theta}: \mathcal{X}_{t}\rightarrow \mathcal{Y}_{t}$, where $\Theta$ represents the parameters of the learner. 

The incremental learner has two goals: to effectively learn the current task (\textit{plasticity}) while retaining performance on all previous tasks (\textit{stability}). This can be accomplished by optimizing the following function where $CE(\cdot)$ represents the Cross-Entropy used in classification, and $Reg(\cdot)$ is a regularization term that penalizes abrupt changes in the neural network weights~\citep{li2017learning,kirkpatrick2017overcoming,rebuffi2017icarl,zhao2020maintaining}:

\vspace{-8pt}
{\small
\begin{align}
\label{equ:train}
\mathcal{L} = CE(f(x_{i,t}), y_{i, t}) + \lambda \cdot Reg(\Theta)
\end{align}}

\subsection{Base Models for AdaCL}
AdaCL can be combined with many base incremental learners. We experimented with four popular, well-established techniques: EWC~\citep{kirkpatrick2017overcoming}, LwF~\citep{li2017learning} iCaRL~\citep{rebuffi2017icarl} and WA~\citep{zhao2020maintaining}. We select baselines that complement each other and serve as strong baselines within the field of incremental learning (Table \ref{tab:baselines}).  

\begin{table}[h]
\caption{Selected models to evaluate the impact of adaptivity in Class-Incremental Learning.}
\label{tab:baselines}
\centering
\small
\begin{tabularx}{\columnwidth}{>{\centering\arraybackslash}m{1 cm}*{4}{>{\centering\arraybackslash}X}}
\hline
method & prior-based & distillation-based & exemplar collection & classifier correction \\ \hline
EWC    &                \ding{51}            &                                   &                     &                       \\
LwF    &                            &           \ding{51}                        &                     &                       \\
iCaRL  &                            &                \ding{51}                   &          \ding{51}           &                       \\
WA     &                            &              \ding{51}                     &         \ding{51}            &      \ding{51}                 \\ \hline
\end{tabularx}
\end{table}

\textbf{EWC.} Elastic Weight Consolidation~\citep{kirkpatrick2017overcoming} is a weighted regularization approach. The authors argue that not all weights contribute equally to learning a new task and estimate the importance of each weight in minimizing the classification loss for the current task: $Reg(\Theta) = || \mathcal{F}(\Theta - \Theta^{\prime}) ||$, where $\Theta^{\prime}$ is the model weights from the previous learning step, $\mathcal{F}$ is the Fisher matrix of the same size as the weight matrices $\Theta$, re-weighting the contributions of each weight to stabilize the important neurons per task.

\textbf{LwF.} Learning-without-Forgetting~\citep{li2017learning} is a knowledge-distillation approach where the teacher branch is the model from the previous task, and the student branch is the current model. The aim is to match the activations of the teacher and student branches, either at the feature or logit level. Formally, LwF minimizes the following objective where $f^{\prime}$ is the model from the previous learning step, and $KL(p_1, p_2)$ is the KL-divergence between two probability distributions $p_{1}$ and $p_{2}$:

\vspace{-10pt}
\begin{align}
\label{KD}
Reg(\Theta) = KL(f(x_{i,t}) , f^{\prime}(x_{i,t}))
\end{align}
\vspace{-10pt}

\textbf{iCaRL.} The Incremental Classifier and Representation Learning~\citep{rebuffi2017icarl} leverages a hybrid approach that involves two main components: exemplar-based memory which is carefully selected to maintain representation and a regularization. The exemplar-based memory module retains a subset of exemplar samples from previous tasks, representing important instances that encapsulate the learned knowledge. By utilizing exemplars, iCaRL ensures the model's ability to recognize and classify past instances while discriminating between learned and new classes. The distillation loss as in Eq. \ref{KD} used for regularization, enables knowledge distillation from previous models to guide the learning process for new tasks. This distillation process allows the model to align logits of new classes with already learned classes to mitigate catastrophic forgetting.

\textbf{WA.} Maintaining Discrimination and Fairness in Class Incremental Learning~\citep{zhao2020maintaining}
is a method that consists of two phases: maintaining discrimination and maintaining fairness. The first phase is similar to the previously established method~\citep{rebuffi2017icarl}. Their study demonstrates that knowledge distillation is not sufficient by itself to prevent the model to treat old classes and new classes fairly since there is a high tendency towards new classes in the classifier layer to minimize the Eq \ref{KD}. Therefore, the second stage named Weight Aligning (WA) focuses on maintaining fairness to correct this classifier bias towards new classes. WA showed that it treats all classes fairly, and significantly improves the overall performance.

\vspace{-9pt}
\subsection{Constancy Assumption in Class Incremental Learning} 
The scalar parameter $\lambda$ balances the contribution of the classification and regularization loss functions. A large value of $\lambda$ ensures minimal weight updates, which can sacrifice learning on the current task. Conversely, a small $\lambda$ yields good performance on the current task but may sacrifice performance on previous tasks, exacerbating catastrophic forgetting. Similarly, requirement for a fixed or predetermined memory size per task may not always be optimal, as it depends on the new task and its relationship to previous tasks. Specifically,where the new task is highly similar to previous tasks, it is possible to retain past knowledge by storing only a small number of representative samples. Conversely, when the new task is significantly distinct, it is reasonable to store a larger number of examples in memory to prevent catastrophic forgetting while learning new tasks.
However, as a common practice, important hyperparameters such as learning rate ($\eta$), regularization strength ($\lambda$), and memory size ($m$) are set to a fixed or pre-defined scalar value throughout all incremental learning sessions with $t \in \mathcal{T}_{1:t}$; such that $\eta_{t} = \eta_{t-1}$, $\lambda_{t} = \lambda_{t-1}$ or $\lambda_{t} = \frac{t*c}{(t*c) + c}$ where c is the number of classes per task. Similarly, $m_{t} = m_{t-1}$ or $m_{t} = \frac{M}{t}$ where M is the pre-defined total memory size.

We hypothesize that the assumption of \textit{constant or pre-defined learning rate, regularization strength, and exemplar size per task} is suboptimal for building accurate lifelong learning machines. Our reasoning is two-fold:

\textbf{Low Plasticity and High Stability.} The incremental learner may encounter a novel object that is highly familiar with the previously learned tasks. For example, it may encounter the category \textit{dog} after observing many other animal categories, such as \textit{{cat, cow, bird}}. In this case, the learner does not need to store many exemplars from previous tasks or to be too plastic, as it can quickly transfer knowledge from the previous tasks where it is similar to the human learning process and referred to \textit{low road transfer} \citep{perkins1992transfer}. Hence, no drastic updates to the learned filters are necessary.

\textbf{High Plasticity and Low Stability.} Conversely, the learner may encounter a novel object that is highly unfamiliar with the previous tasks. For example, it may encounter the category \textit{car} after observing many other animal categories, such as \textit{{cat, cow, bird}}. In this case, the learner would require replaying more exemplars from previous tasks to preserve old knowledge and high plasticity to learn about the novel object with never-before-seen parts, such as wheels.

\vspace{-4pt}
\subsection{AdaCL: Adaptive Continual Learning} 
AdaCL aims to optimize the regularization magnitude $\lambda$ and memory size $m$ as a function of model performance over a set of incremental tasks, conditioned on the current learning task and all previous tasks. We define $\eta(t)={\eta_1, \eta_2, \ldots, \eta_{t-1}, \eta_t}$, and $\lambda(t)~=~{\lambda_1, \lambda_2, \ldots, \lambda_{t-1}, \lambda_t}$, and $m(t)~=~{m_1, m_2, \dots, m_{t-1}, m_t}$ where $\eta_{t}$, $\lambda_{t}$ and $m_{t}$ are predicted by minimizing the following optimization problem:

\vspace{-15pt}
{\small
\begin{align}
\label{equ:optim}
\arg\min_{\eta, \lambda, m} \mathcal{L}(\Theta; V_{t})
= \arg\min_{\eta, \lambda, m} \sum_{i=1}^{|V_{t}|} [CE(f(x_{i,t};\Theta), y_{i,t}) 
\end{align}}

\begin{algorithm}
    \caption{AdaCL: Adaptive Continual Learning}
    \label{algo}
    \small
    \begin{algorithmic}[1]
    \Require
    \Statex{$\theta_{t-1}$} \Comment{model from previous task}
    \Statex{$X_t$} \Comment{dataset from new task}
    \Statex{$M_{t-1} = m_1,\dots,m_{t-2}, m_{t-1}$} \Comment{memory from old tasks}
    \Statex{$V_{t-1} = v_1,\dots,v_{t-2}, v_{t-1}$} \Comment{val. set from tasks seen so far}
    \Statex{$\eta_{space}$}  \Comment{search space for learning rate}
    \Statex{$\lambda_{space}$}  \Comment{search space for regularization}
    \Statex{$m_{space}$}  \Comment{search space for memory}
    \Statex{$configs, epochs$} \Comment{\# of configurations and epochs}
    \State $V_{t} = V_{t-1} \cup v_t \gets X_t$
    \For{$c = 1,\dots,configs$}
    \State $\eta_t \gets \eta_{space}$ \Comment{$\eta$ for new task}
    \State $\lambda_t \gets \lambda_{space}$ \Comment{$\lambda$ for new task}
    \State {$M_{t}: m_{t} \gets m_{space}$} \Comment{memory with a size of $m_t$}
    \State{$D = X_t \cup M_{t-1} \cup M_t$} \Comment{concat new data and memory}
    \For{$e = 1,\dots,epochs$}
    \State{Train  Eq. \ref{equ:train} with $\theta_{t-1}$ and $D$}
    \State{Evaluate Eq. \ref{equ:optim} with $V_t$}
    \EndFor
    \EndFor
    \State \textbf{return} $\theta_{t}, V_t, \eta_t^*, \lambda_t^*,  M_t^*$ \Comment{new model with optimal learning rate, regularization strength and memory size}
    \end{algorithmic}
\end{algorithm}

\noindent Here, $V_{t}$ is a randomly selected class-balanced subset of the current task and previous tasks that guide the model's adaptation with careful consideration of both new and previous tasks' characteristics and prevents bias over certain classes. $\mathcal{L}(\Theta; V_{t})$ is the loss function where the learning rate $\eta$, the regularization coefficient $\lambda$, and memory size per task $m$ is determined by solving the optimization problem. Our adaptive approach, AdaCL (Algorithm \ref{algo}), starts after the first task since it is just a standard batch learning. 

In the following tasks, it retains the model $\theta_{t-1}$ trained on the previous task, receives current task data $X_{t}$, and creates a validation set $V_{t}$. Then, training data $D$ is constructed and trained with Eq. \ref{equ:train} after the configuration for {$\eta_t$,} $\lambda_t$ and $m_{t}$ is selected by Bayesian Optimization (see Section 3.4). After each epoch, the selected configuration is evaluated on the validation set $V_{t}$ with Eq. \ref{equ:optim}. Subsequently, this process is repeated until reaching the total number of configurations. The optimal learning rate $\eta_t^*$, lambda $\lambda_t^*$, and memory size per task $m_t^*$ are determined based on the validation performance.

This approach allows us to automatically adjust the {learning rate, regularization strength, and memory size per task according to the specific learning task based on the given loss function which lets the model find the degree of difficulty itself, avoiding the unrealistic assumption of a fixed learning rate, regularization strength, and memory size throughout the learning process.

\vspace{-5pt}
\subsection{Bayesian Optimization via Parzen Estimator}

We optimize the objective function using multivariate tree-structured parzen estimators (TPE)~\citep{bergstra2011algorithms}. TPE builds a conditional probability tree that maps hyperparameters to their respective model performances. Then it can be used to guide a search algorithm to find the optimal set of hyperparameters for the given model. In this study, TPE is utilized as a search algorithm where it searches within the provided range for learning rate, regularization strength, and memory size per task and then searches for the best value by evaluating across accumulated validation set which consists of previous and new tasks throughout incremental learning sessions. We use Optuna \citep{optuna} for TPE implementation.

\section{Experimental Protocol}
\label{sec:setup}

In this section, we describe our experimental setup, present our findings and results, and provide an ablation study.
\textbf{Datasets.} In this paper, we experiment with \textbf{CIFAR100}~\citep{krizhevsky2009learning} and \textbf{MiniImageNet}~\citep{vinyals}. Each dataset exhibits objects from $100$ different categories. We train all the models with $10$ tasks, with $10$ classes within each learning task on both CIFAR100 and MiniImageNet. Both datasets have 5000 training, and 1000 testing color images per learning task, each with $32\times32$ and $64\times64$ resolution for CIFAR100 and MiniImageNet respectively.

\paragraph{Metrics.}  We resort to the standard metrics for evaluation, accuracy (ACC) which measures the final accuracy averaged over all tasks,  and backward transfer (BWT) which measures the average accuracy change of each task after learning new tasks:

\vspace{-18pt}
{\footnotesize
\begin{align}
ACC=\frac{1}{T}\sum\nolimits_{i=1}^T A_{T,i}
\end{align}}

\vspace{-15pt}
{\footnotesize
\begin{align}
\label{equ:main}
BWT=\frac{1}{T-1}\sum\nolimits_{i=1}^{T-1} (A_{T,i}-A_{i,i})
\end{align}}

\noindent where $A_{T,i}$ represents the testing accuracy of task $T$ after learning task $i$.

\paragraph{Baselines.} EWC, LwF, iCARL, and WA are our direct baselines since we use them as base models in AdaCL. We also compare our common baseline results with OMDP \citep{omdp}. Finally, we select one recent memory-free approach FeTrIL~\citep{fetril}, and one recent memory-based method PODNet~\citep{podnet} to provide more comprehensive insights.

\paragraph{Implementation Details.} 
We employ adaptive hyperparameter optimization on the methods discussed in section 3.2, and compare them with their fixed (original) versions. For the fixed versions, we use the default {$\eta$,} $\lambda$ and $m$ as defined in PYCIL \citep{zhou2021pycil}.

\begin{figure}[h]
  \centering
  \includegraphics[width=0.44\textwidth]{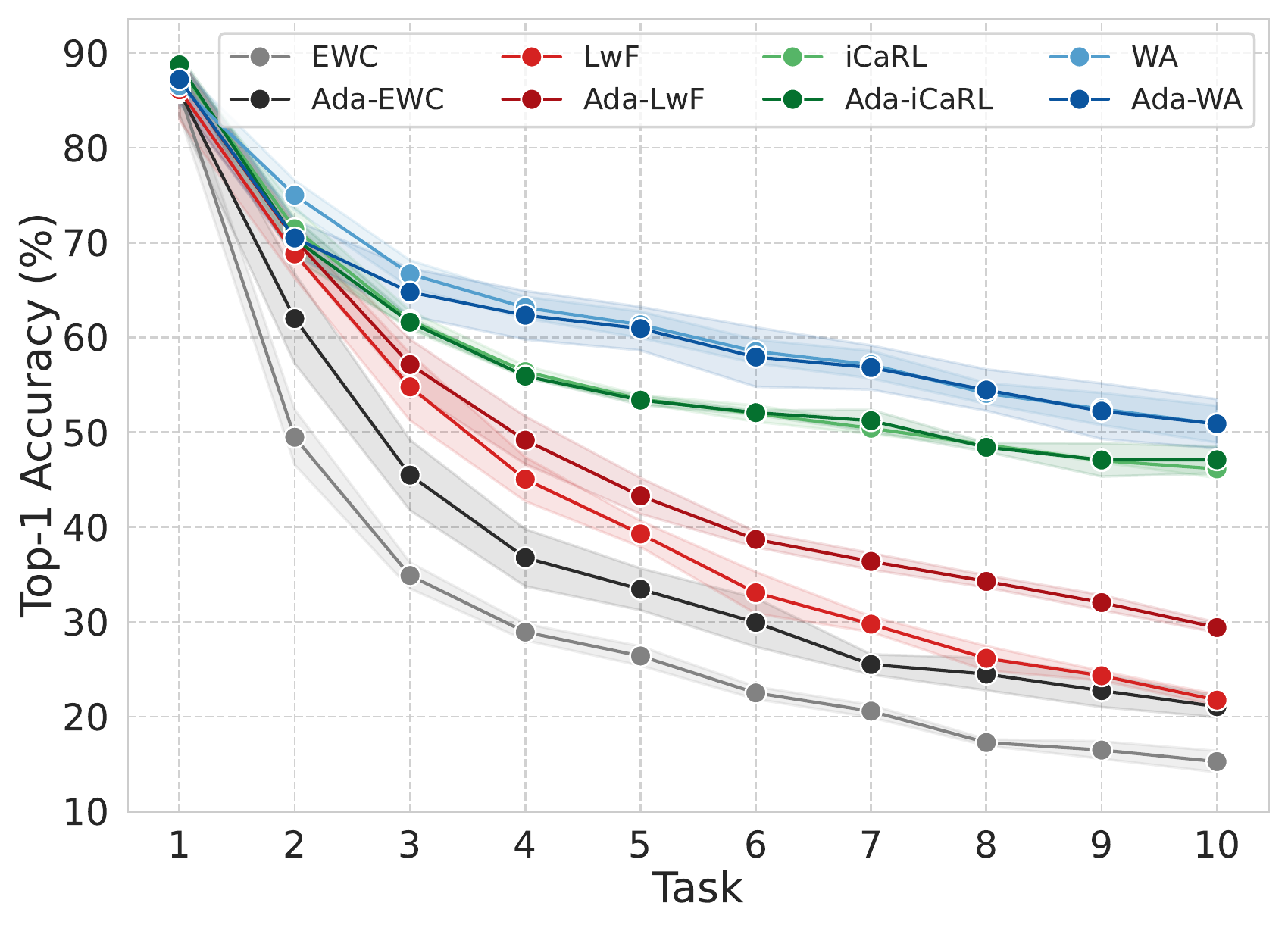}
  \caption{Accuracy after each task on \textbf{CIFAR100}. AdaCL significantly boosts the performance on regularization-based methods and improves the efficiency by storing fewer exemplars on memory-based methods while yielding on par performance.}
  \label{fig:all_cifar100}
\end{figure}

\begin{figure}[h]
  \centering
  \includegraphics[width=0.44\textwidth]{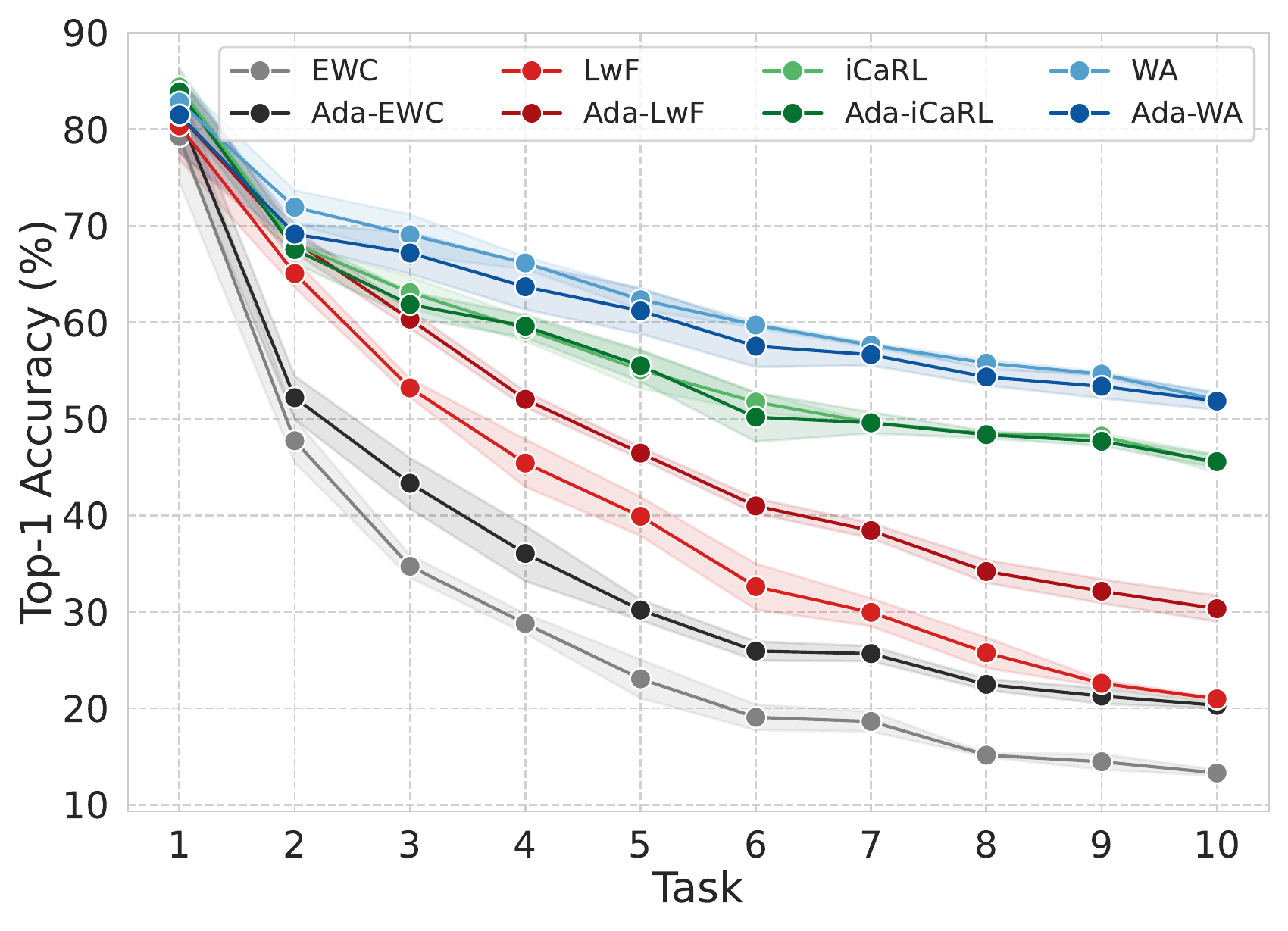}
  \caption{Accuracy after each task on \textbf{MiniImageNet}. The results align with the observations on CIFAR100.}
  \label{fig:all_mini}
\end{figure}

We use ResNet-$32$ as the backbone~\citep{he2016deep}. We set the number of epochs to $100$ but use the Successive Halving~\citep{li2018massively} scheduler for a more efficient search. We use SGD optimizer with momentum parameter set to $0.9$ and weight decay $5e^{-4}$ for the first task and $2e^{-4}$ for the rest of the tasks. The batch size is set to $128$. We run experiments on three different seeds and report their average. We store a small subset of the validation data from each incremental learning step to evaluate the search algorithm. 
The search space for the learning rate and the maximum memory size per class within a task is set to [0.05, 0.1] and 50 respectively. The search space for $\lambda$ is determined based on the ablation experiments and details are given in Appendix \ref{searchspace_ablate}.

\section{Experimental Results}
\paragraph{The Effect of Adaptivity.}
We investigate the efficacy of our adaptive method compared to traditional fixed hyperparameter approaches across the CIFAR100 and MiniImageNet datasets. Our results highlight significant advancements of our adaptive approach AdaCL, particularly notable in regularization-centric techniques like EWC and LwF as illustrated in Figure~\ref{fig:all_cifar100} and Figure~\ref{fig:all_mini}.

\begin{table*}[h]
\caption{Performance comparison of various methods on the CIFAR100 and MiniImageNet datasets in terms of ACC, BWT, and memory size. Baseline methods such as EWC and LwF do not utilize memory. Our proposed methods, denoted with (ours), demonstrate better or competitive performance across both datasets while using less memory.}
\label{tab:results}
\fontsize{5}{6.5}\selectfont
\resizebox{\textwidth}{!}{%
\begin{tabular}{lrrcrrc}
\hline
        & \multicolumn{3}{c}{\textbf{CIFAR100}}         & \multicolumn{3}{c}{\textbf{MiniImageNet}}     \\ \hline
Method & \multicolumn{1}{c}{ACC (\%)} & \multicolumn{1}{c}{BWT (\%)} & Memory Size & \multicolumn{1}{c}{ACC (\%)} & \multicolumn{1}{c}{BWT (\%)} & Memory Size \\ \hline
PODNet    & 39.47 ± 1.39 & -24.14 ± 5.49  & 4500 & 43.49 ± 0.31  & -10.84 ± 10.83 & 4500 \\
OMDP      & 46.94 ± 2.11   &  -28.34 ± 1.28             & 4500 &    46.15 ± 0.55           &  -25.54 ± 4.18              & 4500 \\
FeTrIL    & 27.56 ± 1.50 & -18.92 ± 5.70  & -    & 24.46 ± 1.60  & -15.84 ± 0.73  & -    \\ \hline
EWC       & 15.26 ± 1.37 & -63.96 ± 3.21  & -    & 13.30 ± 0.38  & -60.64 ± 3.11  & -    \\
\textbf{Ada-EWC} \textbf{(ours)}   & 21.06 ± 1.37 & -13.28 ± 3.09  & -    & 20.31 ± 0.39  & -11.86 ± 2.32  & -    \\
LwF       & 21.74 ± 0.73 & -48.88 ± 12.43 & -    & 20.97 ± 0.19  & -50.06 ± 9.59  & -    \\
\textbf{Ada-LwF} \textbf{(ours)}  & 29.41 ± 0.65  & -22.34 ± 4.18  & -    & 30.33 ± 1.65  & -28.71 ± 6.44  & -    \\ \hline
iCaRL     & 46.13 ± 1.35 & -28.84 ± 5.06  & 4500 & 45.83  ± 1.43 & -27.33 ± 5.54  & 4500 \\ 
\textbf{Ada-iCaRL} \textbf{(ours)} & 46.44 ± 2.50 & -28.62 ± 2.85  & 4125 & 46.10 ± 2.26  & -28.77 ± 4.26  & 3950 \\
WA        & 50.84 ± 2.37 & -17.23 ± 1.51  & 4500 & 51.96 ± 0.74  & -22.46 ± 1.67  & 4500 \\
\textbf{Ada-WA} \textbf{(ours)}    & 50.87 ± 3.19 & -20.49 ± 2.88  & 4085 & 51.85 ± 1.12  & -24.22 ± 4.41  & 4050 \\ \hline
\end{tabular}%
}
\vskip -0.1in
\end{table*}

\begin{figure*}[h]
  \centering
  \begin{subfigure}[b]{0.33\textwidth}
    \centering
    \includegraphics[width=\textwidth]{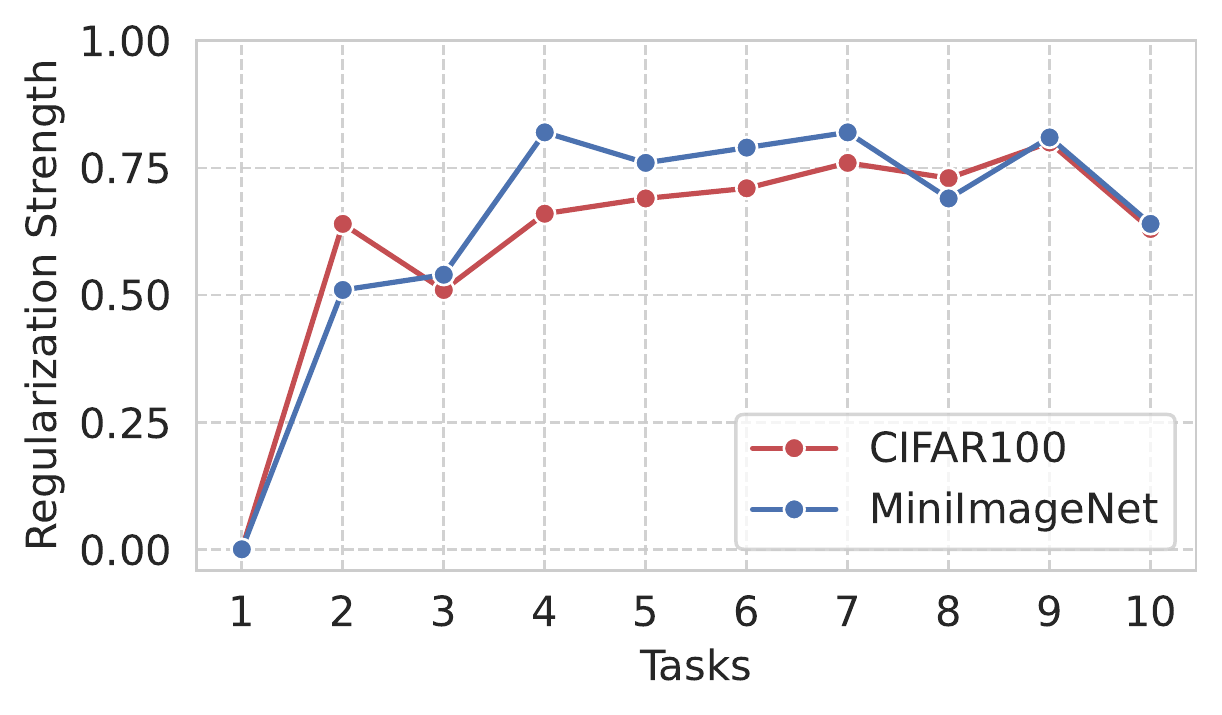}
    \caption{WA regularization strength}
    \label{fig:wa_reg}
  \end{subfigure}
  \begin{subfigure}[b]{0.33\textwidth}
    \centering
    \includegraphics[width=\textwidth]{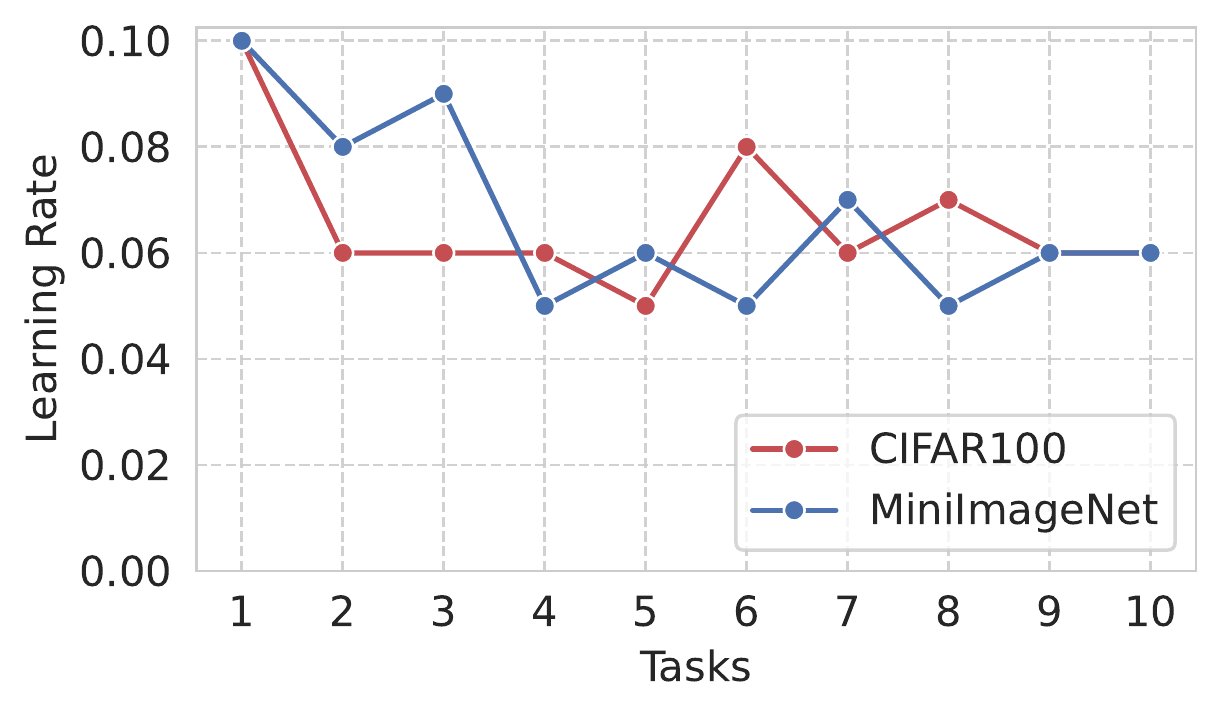}
    \caption{WA learning rate}
    \label{fig:wa_lr}
  \end{subfigure}
  \begin{subfigure}[b]{0.33\textwidth}
    \centering
    \includegraphics[width=\textwidth]{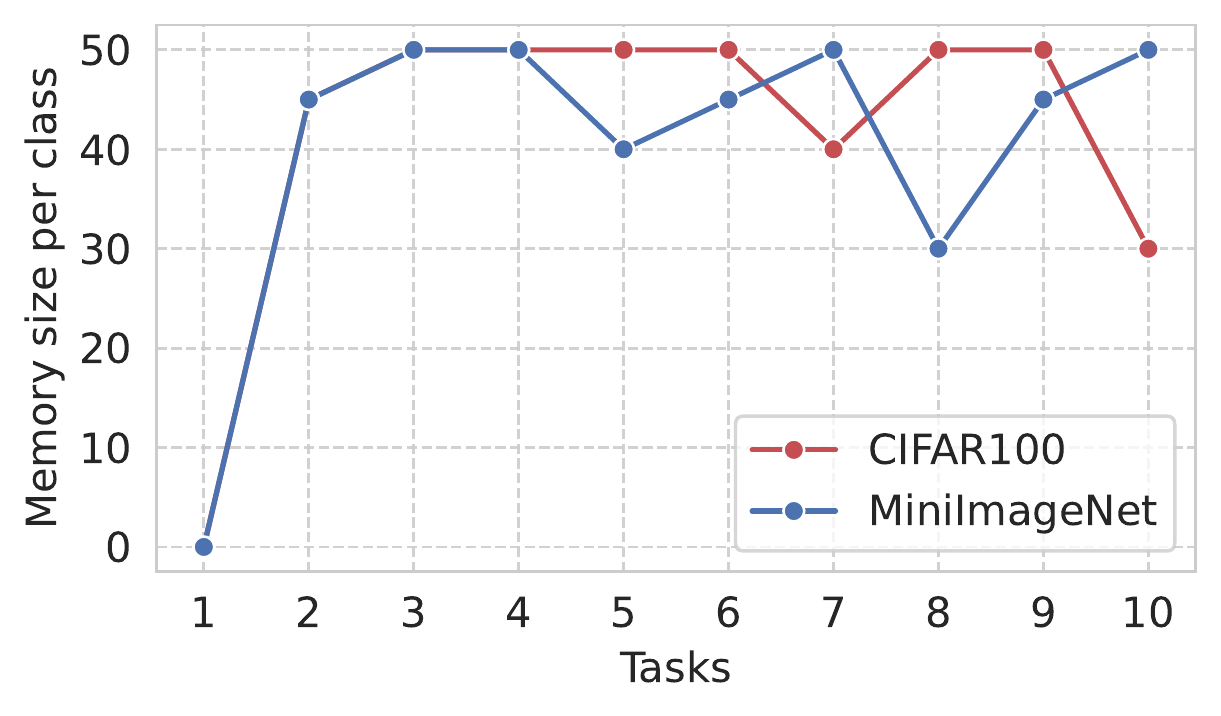}
    \caption{WA memory size}
    \label{fig:wa_mem}
  \end{subfigure}
  \begin{subfigure}[b]{0.33\textwidth}
    \centering
    \includegraphics[width=\textwidth]{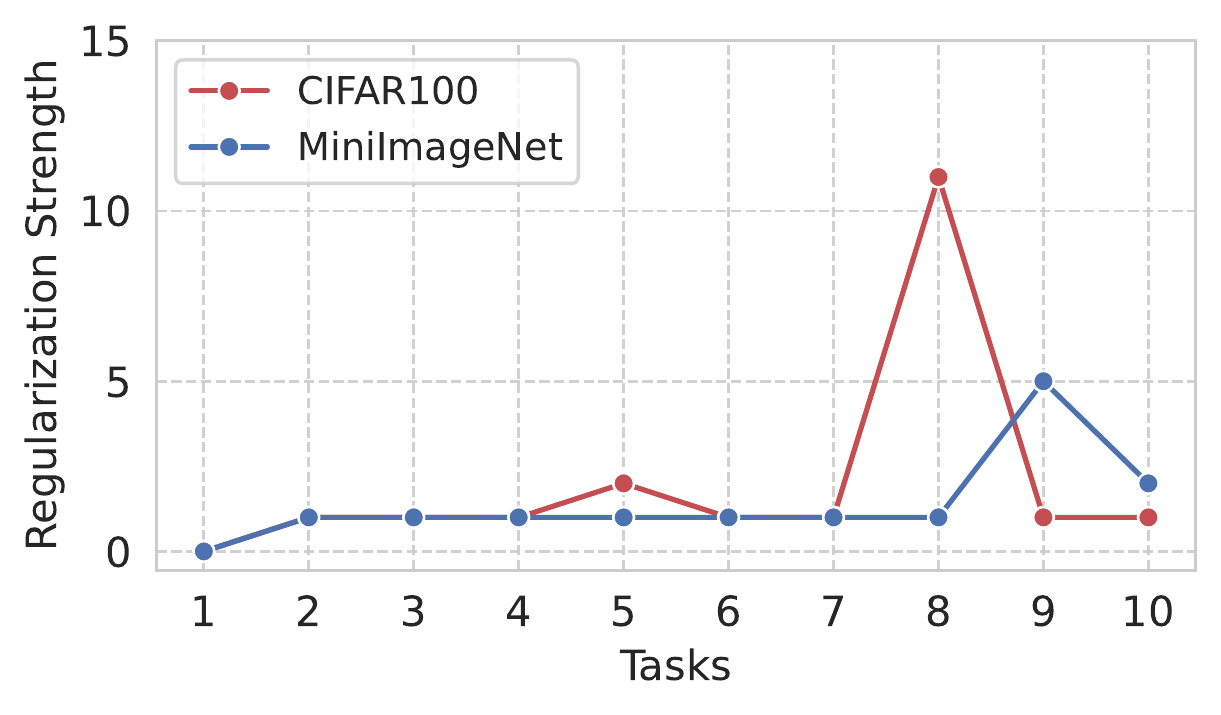}
    \caption{iCaRL regularization strength}
    \label{fig:icarl_reg}
  \end{subfigure}
  \begin{subfigure}[b]{0.33\textwidth}
    \centering
    \includegraphics[width=\textwidth]{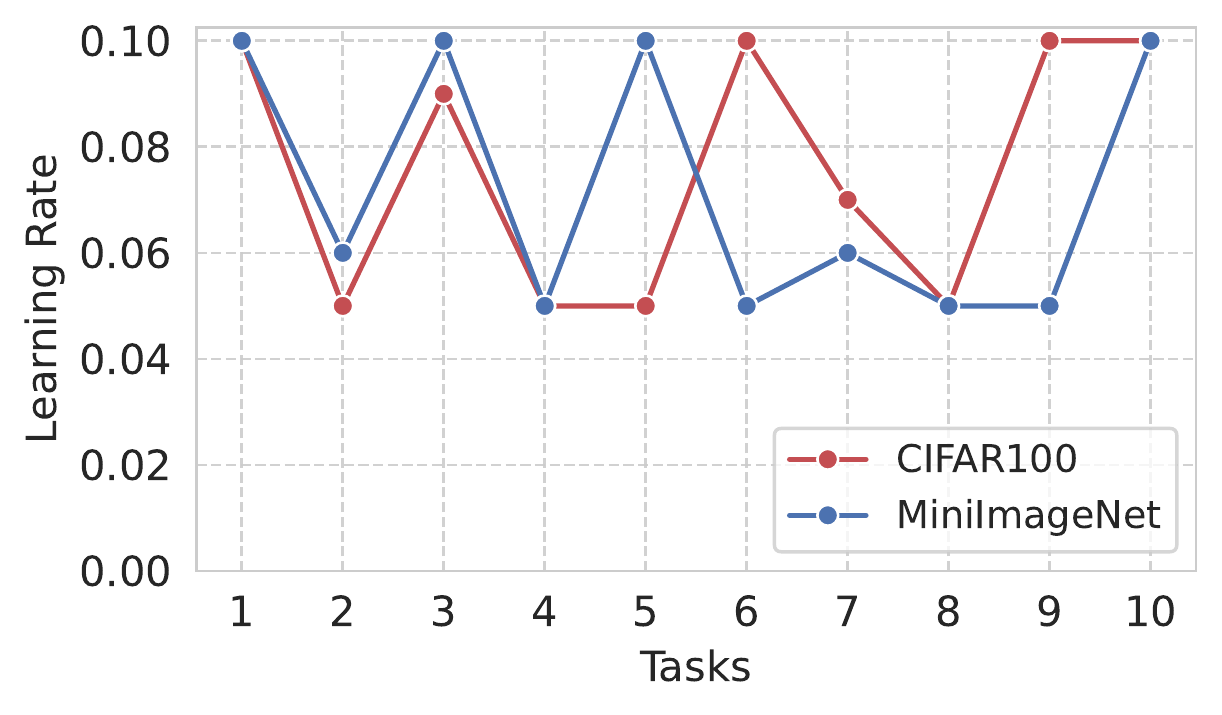}
    \caption{iCaRL learning rate}
    \label{fig:icarl_reg}
  \end{subfigure}
  \begin{subfigure}[b]{0.33\textwidth}
    \centering
    \includegraphics[width=\textwidth]{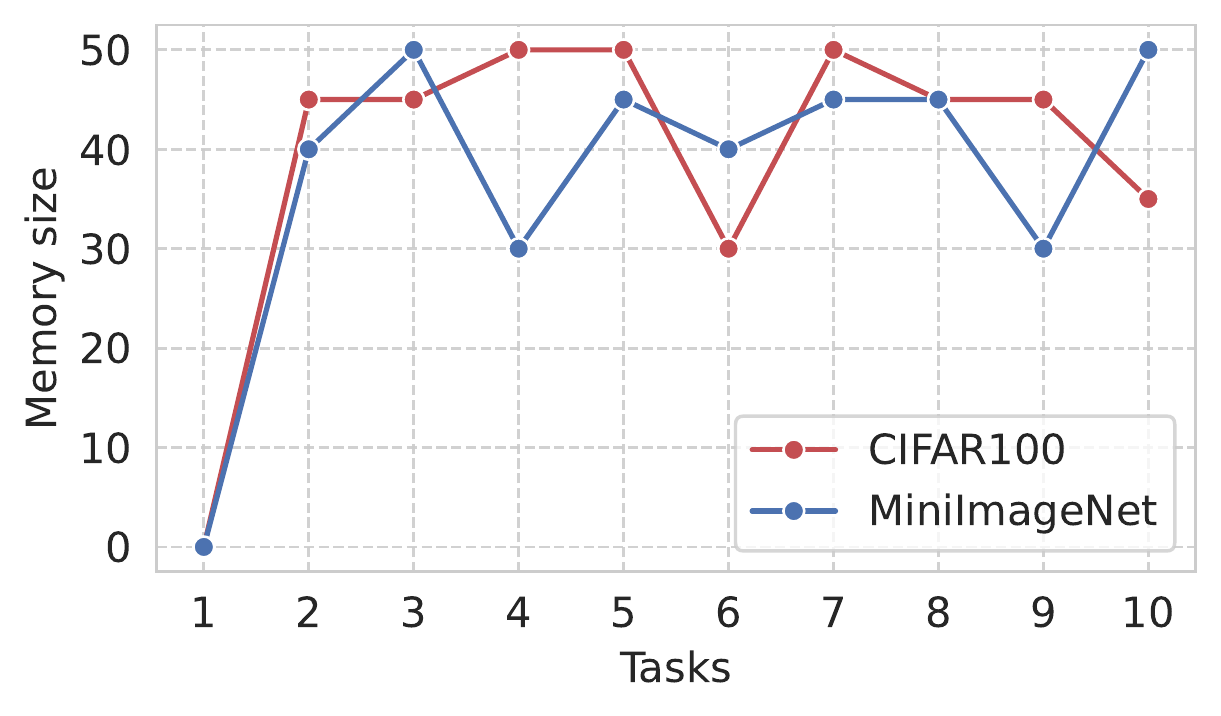}
    \caption{iCaRL memory size}
    \label{fig:icarl_lr}
  \end{subfigure}
  \caption{Adaptive modifications in regularization strength, learning rate, and memory allocation. The selected hyperparameters diversely change across task sequences, datasets, and methods and indicate the necessity of adaptivity in CL.}
  \label{fig:wa_hp}
\vskip -0.1in
\end{figure*}

For example, we find $8\%$ and $10\%$ increase in accuracy while $26\%$ and  $21\%$ improvement in backward transfer on CIFAR100 and MiniImagenet respectively with LwF by adjusting the regularization strength and learning rate. 

In memory-centric methods, we find that they show greater resilience to the changes in hyperparameters. Storing sufficient exemplars aids the model in capturing the distribution of different tasks simultaneously, thereby reducing the reliance on hyperparameter optimization. Despite minor differences, we consistently observe similar accuracy and backward transfer, as seen in Table \ref{tab:results}. 

\paragraph{Comparison with Recent Baselines.} We include results from recent baselines PODNet and FeTrIL to provide comprehensive insights. An intriguing finding is that the initial performance of fixed LwF is outperformed by the recent FeTrIL method, but when we tune LwF, it outperforms FeTrIL by $2\%$ and $6\%$ on CIFAR100 and MiniImageNet, respectively. 
Furthermore, we compare AdaCL with another HPO-based method OMDP, on iCaRL, our only common baseline. Our performance closely aligns with OMDP but a key advantage of our adaptive approach is that it does so while using less memory. 

\vspace{-1pt}
\paragraph{Memory Allocation.}
We investigate the memory allocation of iCaRL and WA by specifically tuning the memory size for these methods. We observe that in our adaptive approach we were able to attain similar results while utilizing less memory compared to those obtained from fixed versions as given in Table \ref{tab:results}.
This is due to AdaCL capability to choose exemplars from both decision boundaries and the center (Figure \ref{fig:tsne}), highlighting how the adaptive approach can achieve comparable results.

\begin{figure*}[h]  
\centering
  \begin{subfigure}[b]{0.27\textwidth}
    \centering
    \includegraphics[width=\textwidth]{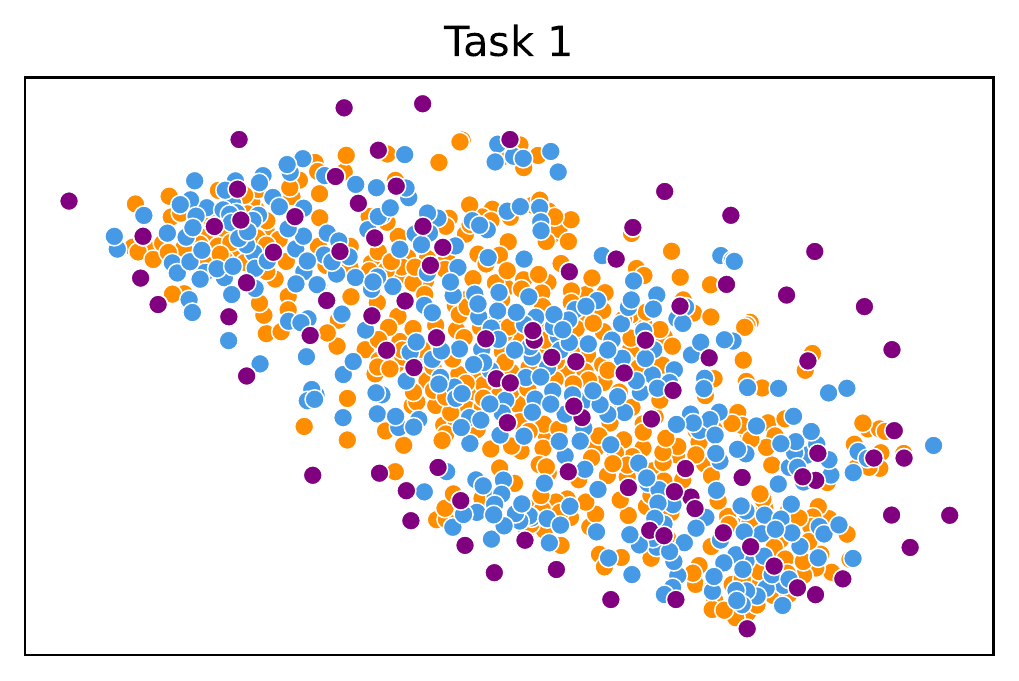}
  \end{subfigure}
  \begin{subfigure}[b]{0.27\textwidth}
    \centering
    \includegraphics[width=\textwidth]{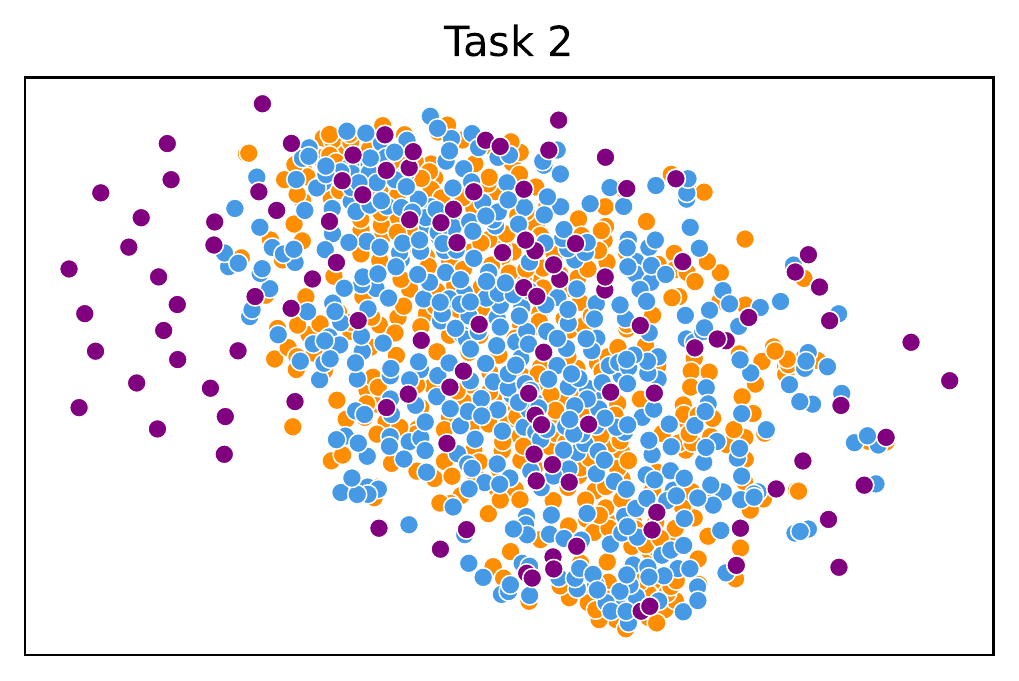}
  \end{subfigure}
  \begin{subfigure}[b]{0.27\textwidth}
    \centering
    \includegraphics[width=\textwidth]{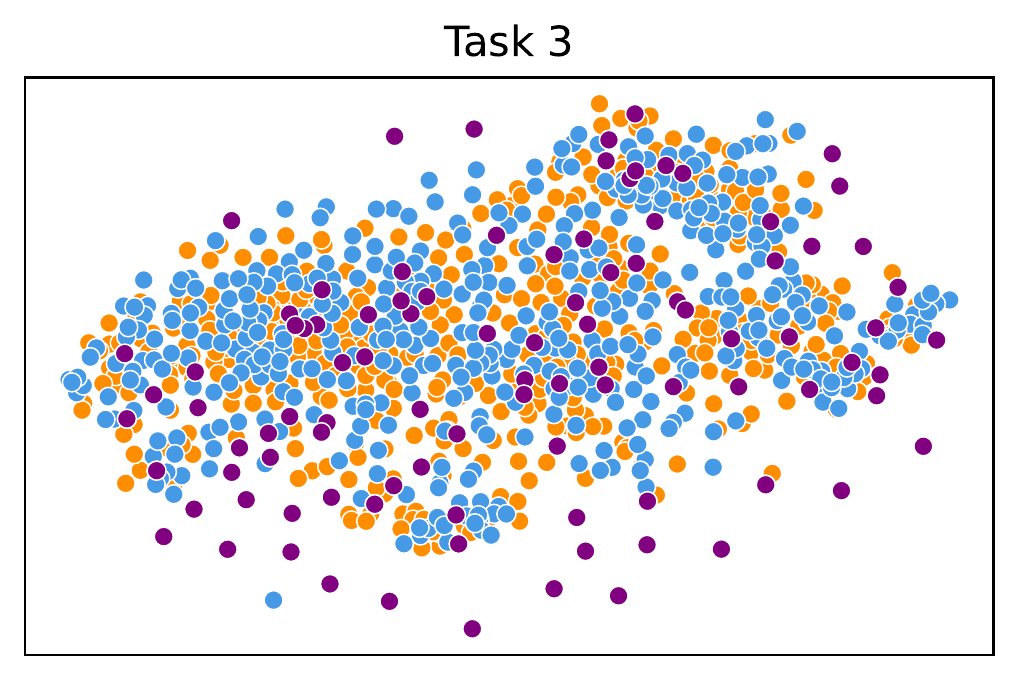}
  \end{subfigure}
   \begin{subfigure}[b]{0.27\textwidth}
    \centering
    \includegraphics[width=\textwidth]{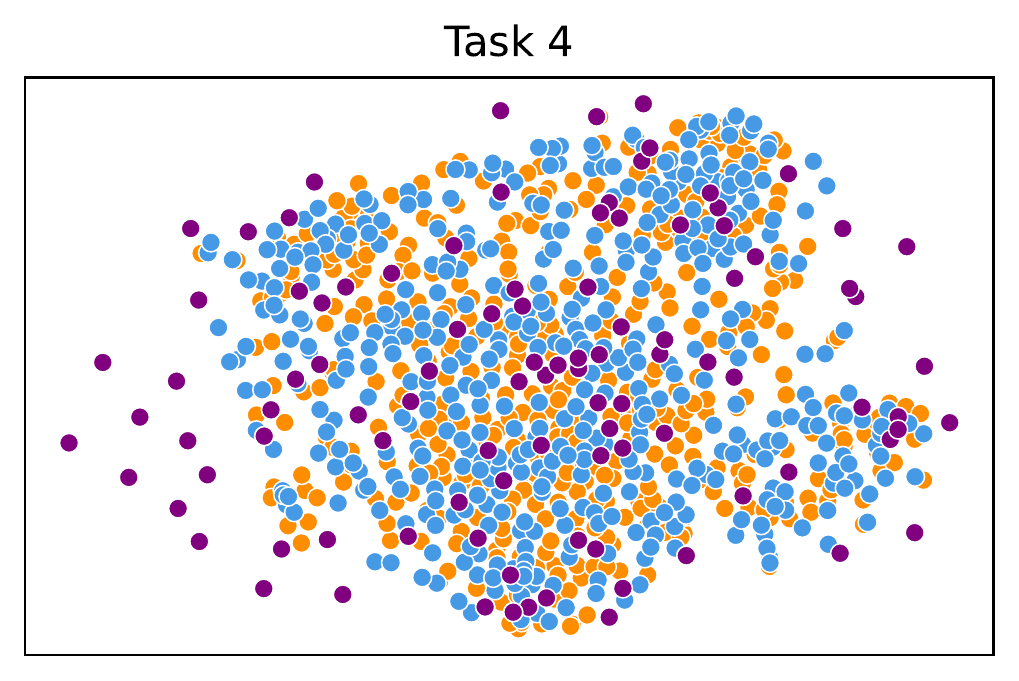}
  \end{subfigure}
   \begin{subfigure}[b]{0.27\textwidth}
    \centering
    \includegraphics[width=\textwidth]{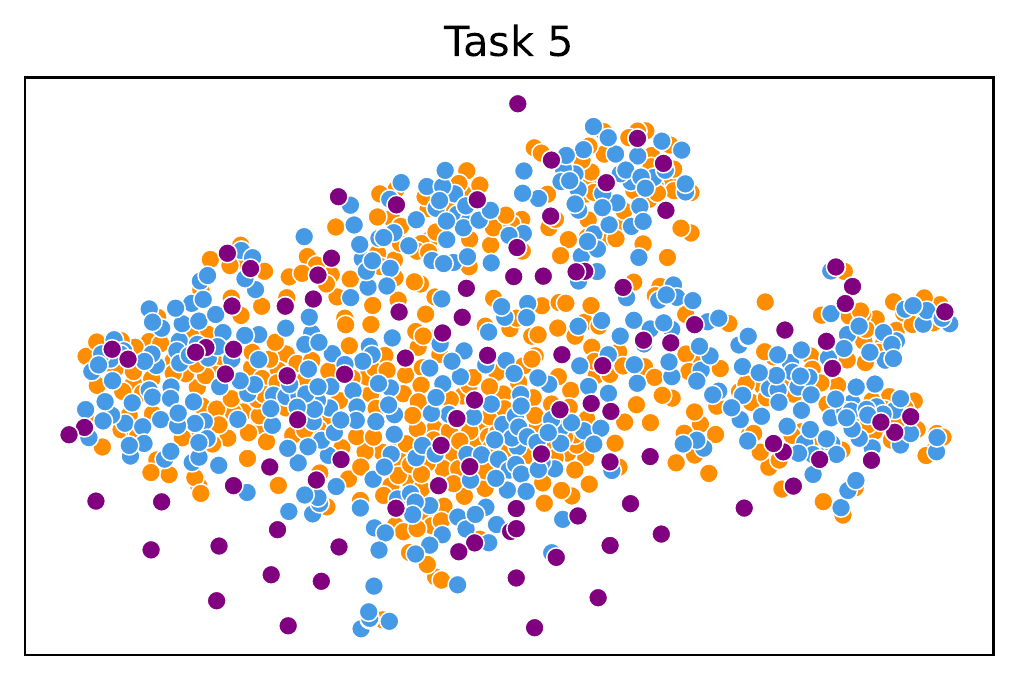}
  \end{subfigure}
   \begin{subfigure}[b]{0.27\textwidth}
    \centering
    \includegraphics[width=\textwidth]{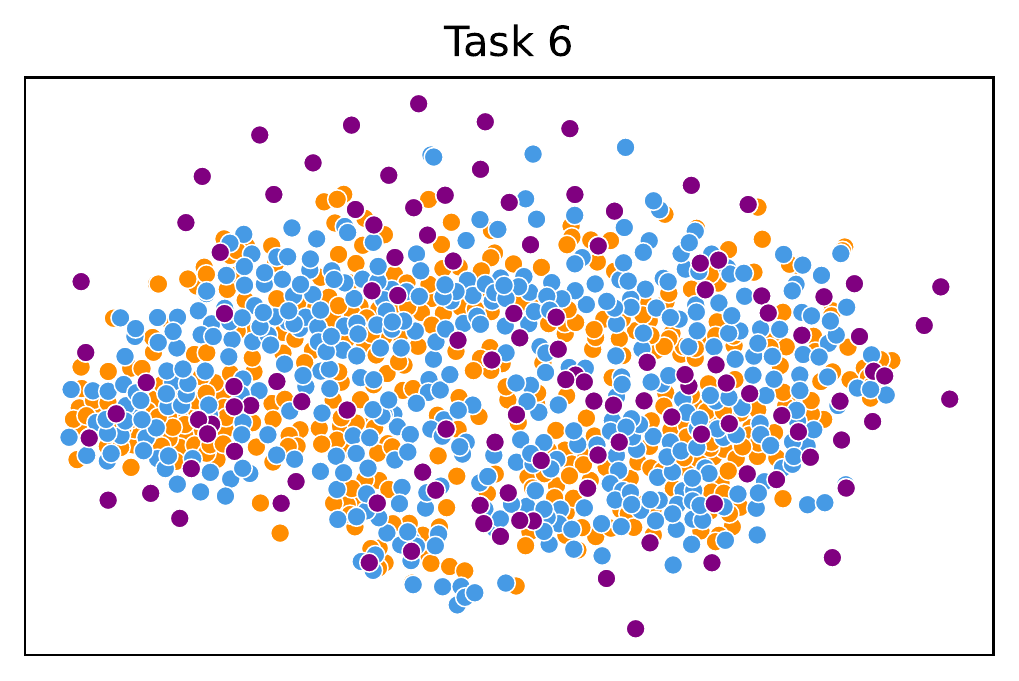}
  \end{subfigure}
   \begin{subfigure}[b]{0.27\textwidth}
    \centering
    \includegraphics[width=\textwidth]{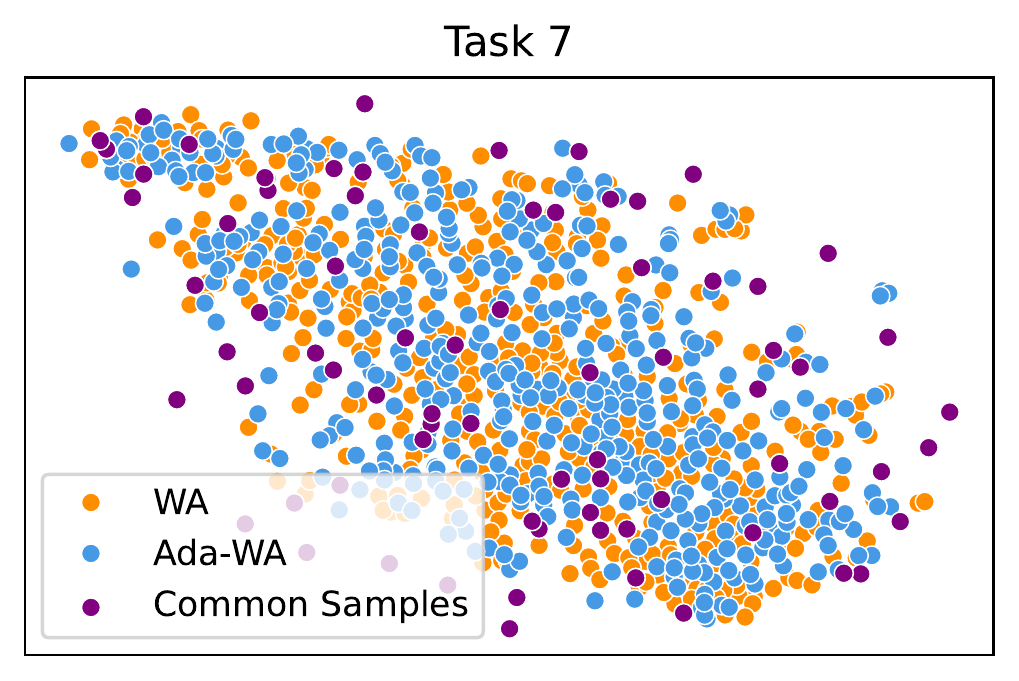}
  \end{subfigure}
   \begin{subfigure}[b]{0.27\textwidth}
    \centering
    \includegraphics[width=\textwidth]{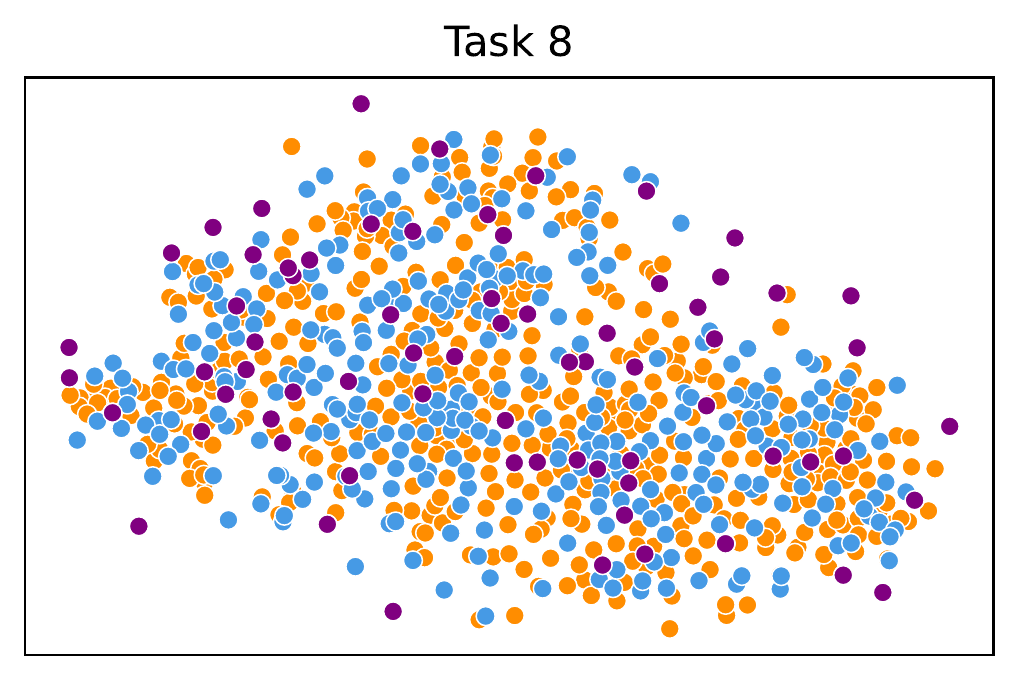}
  \end{subfigure}
  \begin{subfigure}[b]{0.27\textwidth}
    \centering
    \includegraphics[width=\textwidth]{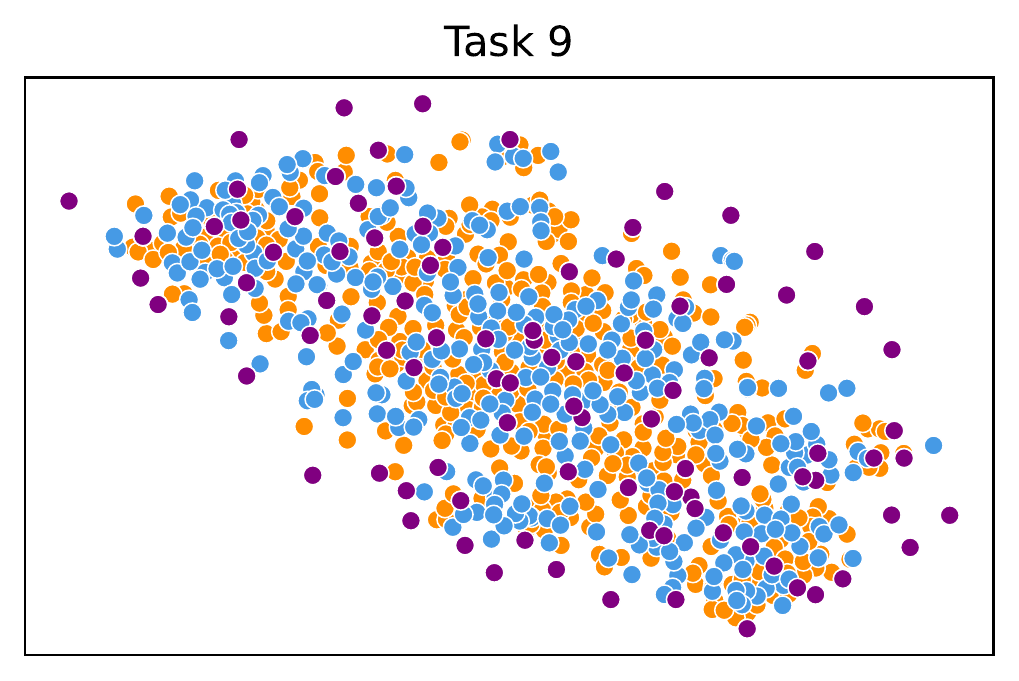}
  \end{subfigure}
  \caption{t-SNE plots of selected exemplars. Ada-WA selects exemplars from boundaries and center. This way, it is able to achieve on-par performance with less memory. The final task is omitted from the visualization, since memory selection is not necessary for it.}
  \label{fig:tsne}
  \vskip -0.1cm
\end{figure*}

\paragraph{Exploring Hyperparameter Dynamics.}
We observe the selected hyperparameters throughout the process of continual learning and reveal intriguing dynamics in the adjustment of regularization strength, learning rate, and memory size across tasks. This observation highlights the crucial role of adaptability in continual learners, allowing them to adapt dynamically to the changing demands of each task. 

For example, Figure \ref{fig:wa_hp} presents the chosen hyperparameters for WA and iCaRL, illustrating the subtle adjustments made throughout the learning. Please refer to Appendix \ref{hpo_dynamics} for other methods.

\vspace{-3pt}
\subsection{Ablation Study}
In Table 3, we provide a comprehensive analysis of how different hyperparameters interact with model performance. Our findings reveal that the performance of the model is intricately linked to the interplay of various hyperparameters. While optimizing solely the learning rate and regularization strength yields the highest accuracy, we also incorporate with memory size to enhance memory allocation efficiency.

Our comprehensive hyperparameter optimization strategy showcases an enhancement in memory efficiency by minimizing the amount of stored exemplars while maintaining on-par accuracy. This insight underscores the importance of not only optimizing individual hyperparameters but also understanding their collective impact on model performance, particularly in scenarios where resource constraints necessitate efficient memory allocation in practical applications.

\vspace{3pt}
\begin{table}[h]
\caption{The findings of our ablation study on the WA method, where different hyperparameters were tuned simultaneously and individually on CIFAR100 dataset. Tuning all hyperparameters simultaneously results in a negligible decrease in ACC but yields improvements in memory.}
\fontsize{12}{18}\selectfont
\label{tab:wa_ablate}
\resizebox{\columnwidth}{!}{%
\begin{tabular}{ccccc}
\hline
\textbf{Regularization} &\textbf{ Learning} & \textbf{Memory} & \textbf{ACC}  & \textbf{Stored Memory} \\
\textbf{Strength}       & \textbf{Rate}     & \textbf{Size}   & \textbf{(\%)}       & \textbf{Size}          \\ \hline
    -           &      -    &    -    & 53.48    & 4500          \\
\checkmark              &    -      &     -   & 54.00    & 4500          \\
     -          & \checkmark         &    -    & 53.65    & 4500          \\
      -         &      -    & \checkmark       & 51.02    & 3350          \\
\checkmark               & \checkmark         &     -   & 54.24    & 4500          \\
\checkmark               &    -      & \checkmark       & 51.73    & 3350          \\
     -          & \checkmark         & \checkmark       & 51.40    & 3750          \\
\checkmark               & \checkmark         & \checkmark       & 53.16    & 4250          \\ \hline
\end{tabular}%
}
\end{table}

Finally, in Table \ref{tab:random_memory}, we investigate the impact of employing random memory selection as an alternative to herding under the WA method. Although there is a slight tendency to decrease in both incremental accuracy and forgetting, we found that the adoption of random memory selection leads to an insignificant change in accuracy and forgetting.

\begin{table}[h]
\caption{An investigation into the WA method when using random and herding as the memory selection strategies on CIFAR100 dataset. Findings indicate that random memory selection produces similar results to herding.}
\fontsize{5}{10}\selectfont
\label{tab:random_memory}
\resizebox{\columnwidth}{!}{%
\begin{tabular}{llcc}
\hline
\textbf{Method}        & \textbf{selector} &\textbf{ ACC (\%) }    & \textbf{BWT (\%)}      \\ \hline
WA            & herding & 50.84 ± 2.37 & -17.23 ± 1.51 \\
Ada-WA        & herding & 50.87 ± 3.19 & -20.49 ± 2.88 \\
WA            & random  & 49.95 ± 2.72 & -17.59 ± 2.22 \\
Ada-WA        & random  & 49.96 ± 2.55 & -21.51 ± 1.28 \\ \hline
\end{tabular}%
}
\end{table}

\newpage
\section{Conclusion}
This study introduces the idea of adaptive hyperparameter tuning for Class-Incremental Learning. These hyperparameters are treated as tunable variables that can be adjusted for an each new task according to the learner's current condition and the complexity of the task. Leveraging the sample-efficiency of Bayesian Optimization, the paper presents a methodology to predict the optimal values for these hyperparameters in each learning task. By conducting experiments on well-established benchmarks, the study showcases the remarkable enhancements in performance achieved through adaptive learning, resulting in improved accuracy, diminished forgetting, and less memory. Potential avenues for future investigation could involve the reduction of hyperparameter tuning costs (e.g. via warm-starting) and the exploration of alternative methods for constructing or optimizing the validation set.

To sum up, our study leads the way in introducing the concept of adaptive hyperparameter optimization in Class-Incremental Learning, with a mindful consideration of the limitations we've recognized. As the field further advances, we anticipate that these insights will shape the evolution of advanced continual learning approaches, empowering deep neural networks to adapt to streams of real-world tasks.

\section*{Acknowledgements}
This work is supported by; TAILOR, a project funded by the EU Horizon 2020 research and innovation programme under GA No. 952215, and the Dutch national e-infrastructure with the support of SURF, Cooperative using grant no. EINF-4569, and the Turkish MoNE scholarship.

\bibliography{biblo}

\vfill
\clearpage
\appendix
\section{Appendix}

\subsection{Search Space Range}
\label{searchspace_ablate}
In this section, we present the findings of our search space ablation study conducted to determine the optimal range of regularization strength. Our analysis revealed that regularization strengths within the interval of [1, 100] yielded optimal performance for LwF, thus it is used in our main experiments. Similarly, for EWC  the optimal regularization strength falls within the range of [1, 50000] and is employed in our main experiments.

\begin{table}[h]
\caption{Ablation of the search space for regularization hyperparameter on EWC and LwF methods. The search space intervals were determined based on achieving optimal accuracy results. Best results are highlighted in bold.}
\label{tab:reg_searchspace}
\resizebox{\columnwidth}{!}{%
\begin{tabular}{lcccc}
\hline
\multicolumn{1}{c}{Search Space} & {[}1,1000{]} & {[}1,25000{]} & {[}1,50000{]} & {[}1,10000{]} \\ \hline
EWC & 9.17  & 22.02 & \textbf{22.39} & 21.09         \\ \hline
\multicolumn{1}{c}{Search Space} & {[}1,10{]}   & {[}1,25{]}    & {[}1,50{]}    & {[}1,100{]}   \\ \hline
LwF & 23.99 & 25.49 & 26.73          & \textbf{28.9} \\ \hline
\end{tabular}%
}
\end{table}

\subsection{Hyperparameter Dynamics}
\label{hpo_dynamics}
In this section, we present an overview of the selected hyperparameters for the EWC and LWF methods. Our analysis highlights that our adaptive approach allows all models to flexibly adjust their hyperparameters across different tasks as illustrated in Figure \ref{fig:ewc_hp} and \ref{fig:lwf_hp}. This flexibility plays a role in improving the performance of the models over time by adaptively adjusting hyperparameters.
\begin{figure}[h]
  \centering
  \begin{subfigure}[b]{0.32\textwidth}
    \centering
    \includegraphics[width=\textwidth]{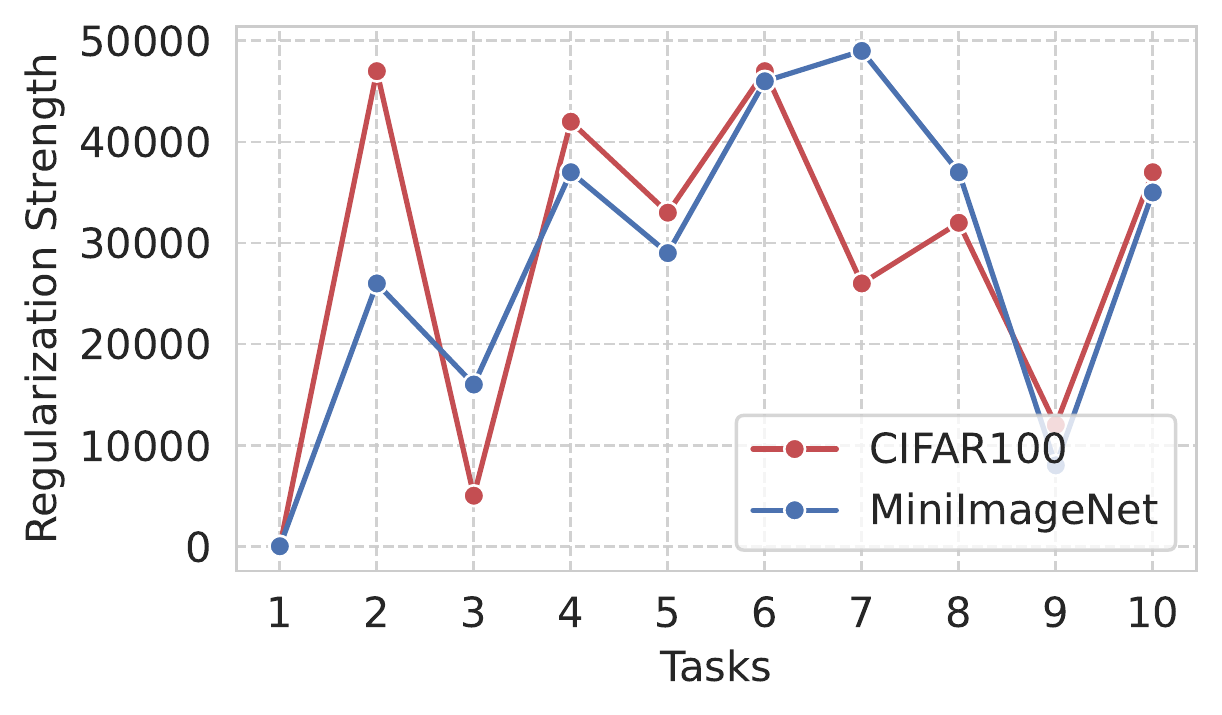}
    \caption{Selected regularization strength}
    \label{fig:ewc_reg}
  \end{subfigure}
  \begin{subfigure}[b]{0.32\textwidth}
    \centering
    \includegraphics[width=\textwidth]{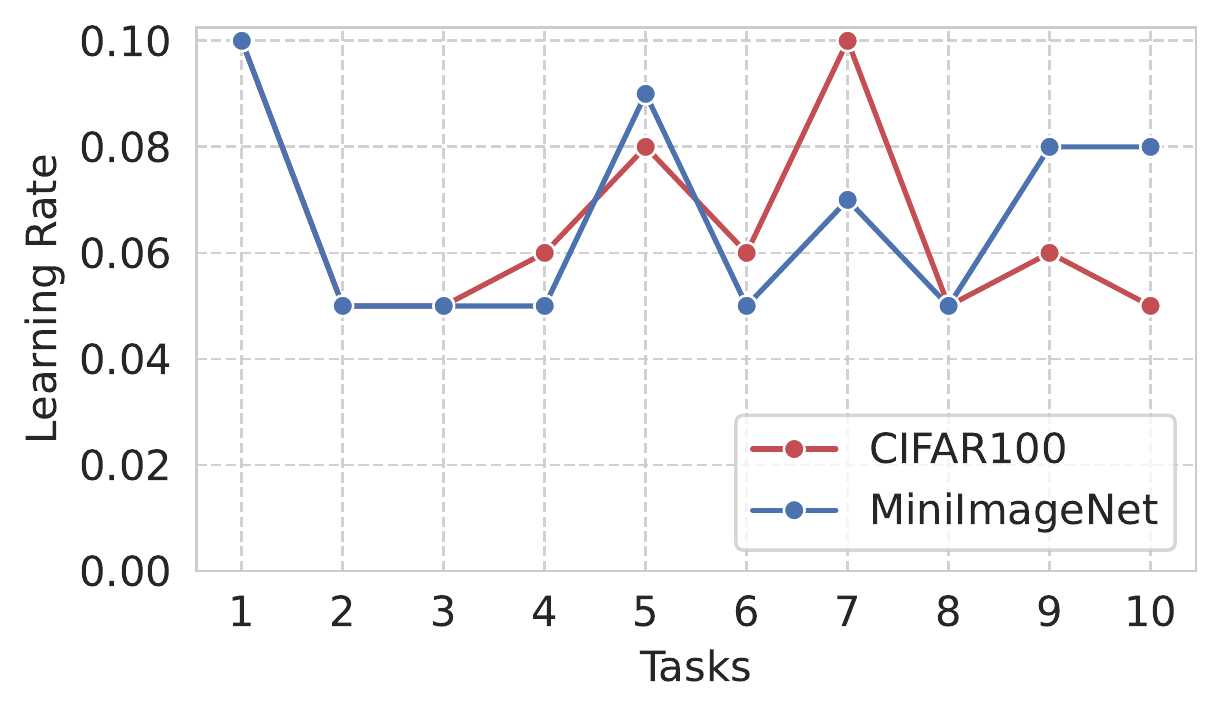}
    \caption{Selected learning rate}
    \label{fig:ewc_lr}
  \end{subfigure}
  \caption{Adaptive adjustments in (a) regularization strength and (b) learning rate for EWC across various task sequences and datasets.}
  \label{fig:ewc_hp}
\end{figure}

\begin{figure}[t]
\vskip -12.5cm
  \centering
  \begin{subfigure}[b]{0.32\textwidth}
    \includegraphics[width=\textwidth]{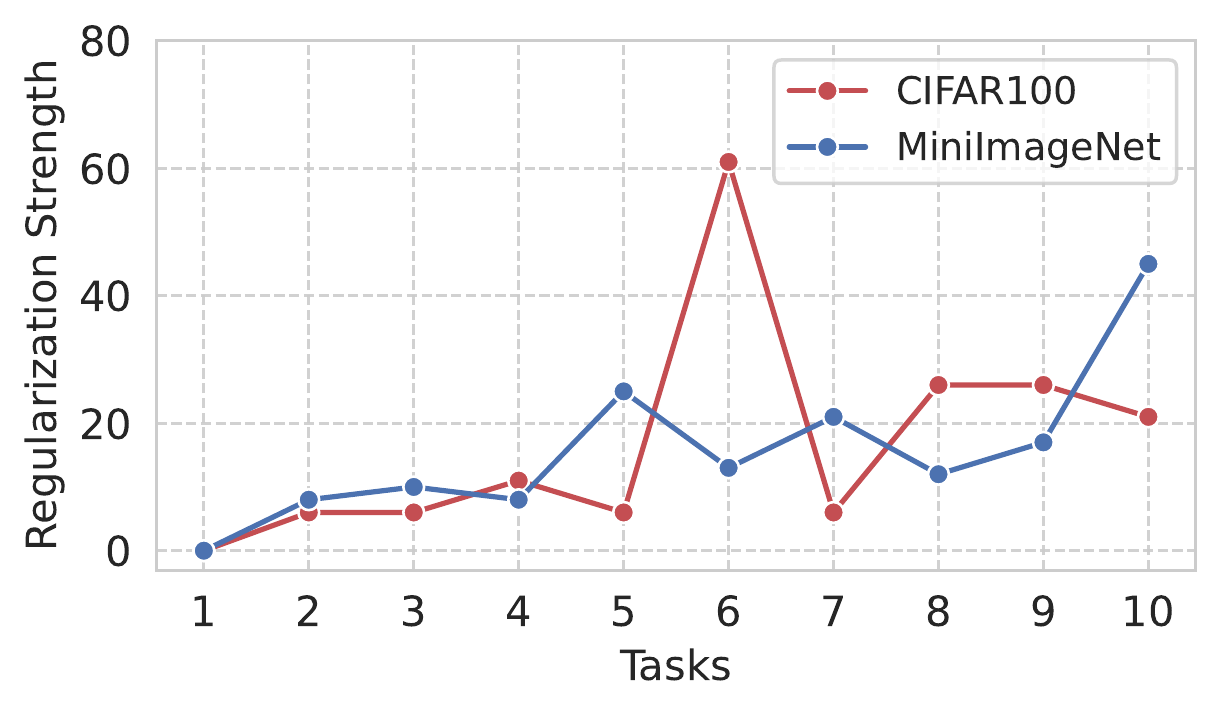}
    \caption{Selected regularization strength}
  \end{subfigure}
  \begin{subfigure}[b]{0.32\textwidth}
    \includegraphics[width=\textwidth]{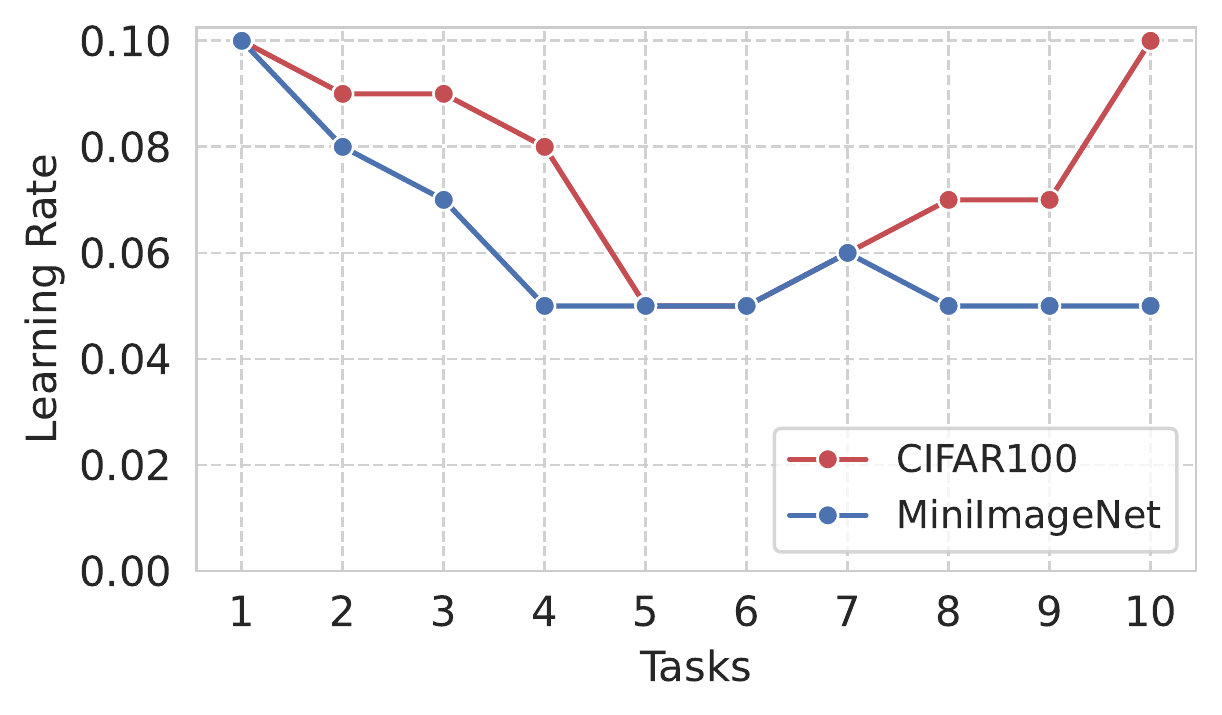}
    \caption{Selected learning rate}
  \end{subfigure}
  \caption{LwF's (a) regularization strength and (b) learning rate change dynamically across different task sequences and datasets.}
  \label{fig:lwf_hp}
\end{figure}

\end{document}